\documentclass{article}
\usepackage[T1]{fontenc} 
\usepackage[utf8]{inputenc} 
\usepackage{ismir,amsmath,cite,url}
\usepackage{graphicx}
\usepackage{color}
\usepackage[bookmarks=false]{hyperref}
\usepackage{lineno}

\title{Towards Estimating Personal Values in Song Lyrics}

\multauthor
{Andrew M. Demetriou$^1$ \hspace{0.3em} Jaehun Kim$^2$ \hspace{0.3em} Sandy Manolios$^1$ \hspace{0.3em} Cynthia C.~S. Liem$^1$ } {\\
  $^1$ Delft University of Technology\\
$^2$ SiriusXM/Pandora\\
{\tt\small a.m.demetriou@tudelft.nl / c.c.s.liem@tudelft.nl }
}

\def\authorname{F. Author, S. Author, and T. Author}

\begin{document}

\maketitle
\begin{abstract}
Most music widely consumed in Western Countries contains song lyrics, with U.S. samples reporting almost all of their song libraries contain lyrics. In parallel, social science theory suggests that personal values - the abstract goals that guide our decisions and behaviors - play an important role in communication: we share what is important to us to coordinate efforts, solve problems and meet challenges. Thus, the values communicated in song lyrics may be similar or different to those of the listener, and by extension affect the listener's reaction to the song. This suggests that working towards automated estimation of values in lyrics may assist in downstream MIR tasks, in particular, personalization. However, as highly subjective text, song lyrics present a challenge in terms of sampling songs to be annotated, annotation methods, and in choosing a method for aggregation. In this project, we take a perspectivist approach, guided by social science theory, to gathering annotations, estimating their quality, and aggregating them. We then compare aggregated ratings to estimates based on pre-trained sentence/word embedding models by employing a validated value dictionary. We discuss conceptually 'fuzzy' solutions to sampling and annotation challenges, promising initial results in annotation quality and in automated estimations, and future directions. 
\end{abstract}

\section{Introduction}\label{sec:introduction}
\begin{figure}
 \centerline{\framebox{
 \includegraphics[width=1\columnwidth]{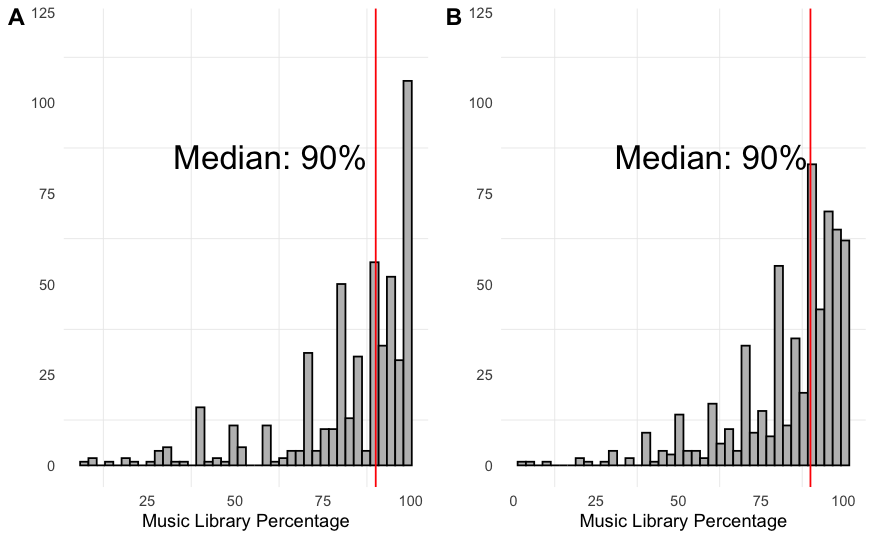}}}
 \caption{Distribution of self-reported percentage of music library containing lyrics from two representative US samples, n=505 and n=600 respectively.}
 \label{fig:lyrics_percentages}
\end{figure}

Popular music in Western countries almost always contains lyrics, making song lyrics a widely, repeatedly consumed \cite{conrad2019extreme} form of text. Over 616 million people subscribe to streaming services worldwide\footnote{\url{https://www.musicbusinessworldwide.com/files/2022/12/f23d5bc086957241e6177f054507e67b.png}}, many of whom stream more than an hour of music every day\footnote{\url{https://www.gwi.com/reports/music-streaming-around-the-world}}. Lyrics have been shown to be a salient component of music \cite{demetriou2018vocals}, and out of over 1400 number-1 singles in the UK charts, only 30 were instrumental\footnote{\url{https://en.wikipedia.org/wiki/List_of_instrumental_number_ones_on_the_UK_Singles_Chart}}. The two representative US population samples that were our annotators indicate a median 90\%\ of songs in their libraries contain lyrics (Figure~\ref{fig:lyrics_percentages}). 

It is thus not surprising that informative relationships between popular songs and their lyrical content have been shown: e.g., country music lyrics rarely include political concepts \cite{van2005world}, and songs with more typical \cite{north2020relationship} and more negative \cite{brand2019cultural} lyrics appear to be more successful. \cite{howlin2020patients} showed that patients are more likely to choose music with lyrics when participating in music-based pain reduction interventions, although melody had an overall larger effect \cite{ali2006songs} showed that lyrics enhance self reported emotional responses to music, and \cite{brattico2011functional} showed a number of additional brain regions were active during the listening of sad music with lyrics, vs.\ sad music without lyrics. In fields closer to MIR, \cite{kim2020butter} show that estimating psychological concepts from lyrics showed a small benefit in a number of MIR tasks, and \cite{preniqi2022more} showed a correlation between moral principles estimated from song lyrics and music preferences. 

A connection between music lyrics and music preferences anticipated by theory involves the personal values perceived in the lyrics by listeners. Prior work has shown correlations between an individual's values, and the music they listen to \cite{manolios2019influence, gardikiotis2012rock, swami2013metalheads, preniqi2022more}, suggesting that we seek music in line with our principles. Yet we have not seen an attempt to measure perceived personal values expressed in the lyrics themselves via human annotation or automated methods. 

In this work we take a first step towards the automated estimating the values perceived in song lyrics. As artistic and expressive language, lyrics are ambiguous text: they contain different forms of analogy and wordplay \cite{sandri2023don}. Thus we take a perspectivist approach to the annotations: because we expect that perceptions will vary substantially more than in other annotation tasks, we aim to represent the general perceptions of only one population. We account for the subjectivity by gathering a large number of ratings (median 27) per song from a targeted population sample (U.S.), of 360 carefully sampled song lyrics, using a psychometric questionnaire that we adjust for this purpose. We treat values in line with theory: as ranked lists, using Robust Ranking Aggregation (RRA) to arrive at our 'ground truth'. We then gather estimates from word embedding models, by measuring semantic similarity between the lyrics and a validated dictionary. We show that ranked lists from estimates correlate moderately with annotation aggregates. We then discuss the implications of our results, the limitations of this project, and anticipated future work. 

\section{Personal Values}
\begin{figure}
 \centerline{\framebox{
 \includegraphics[width=1\columnwidth]{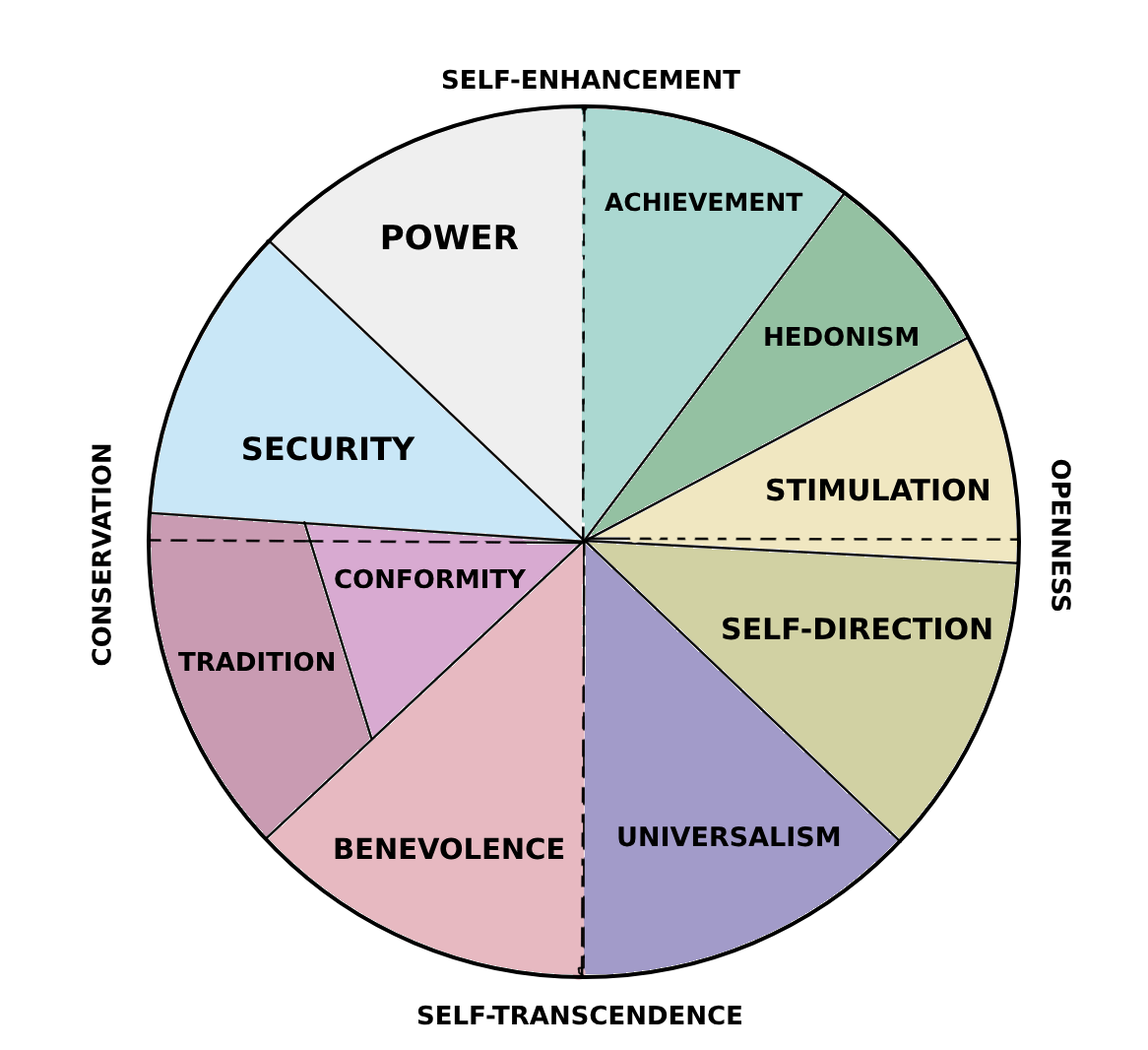}}}
 \caption{Visualization of the Schwartz 10-value inventory from \cite{schwartz1992universals} used in this paper, such that more abstract values of Conservation, vs. Openness to Change, and Self-transcendence vs. Self-enhancement form 4 higher-order abstract values. Illustration adapted from \cite{maio2010mental}. }
 \label{fig:circle}
\end{figure}

The modern study of human values spans over 500 samples in nearly 100 countries over the past 30 years, and has shown a relatively stable structure \cite{sagiv2022personal}, as illustrated in Figure~\ref{fig:circle}. Personal values are a component of personality, defined as the hierarchy of principles that guide a person's thoughts, behaviors, and the way they evaluate events \cite{schwartz1987toward, schwartz2012overview}. Basic human values can be used to describe people or groups: social science theory suggests that each person uses a hierarchical list of values as life-guiding principles \cite{rokeach1973nature}, such that we prioritize some values over others as we make decisions. Schwartz's theory is the most widely used in social and cultural psychology, and has shown correlations with important behaviors, ranging from political affiliation to personal preferences \cite{sagiv2022personal}.

We communicate our values in order to gain cooperation and coordinate our efforts, according to Schwartz \cite{schwartz1992universals}. Thus our values will manifest in the words that we use~\cite{boyd2017language}. Although personal values are traditionally measured by having individual people complete validated psychological questionnaires, it has been argued that values may be clearly expressed in the speech and text that we produce \cite{boyd2017language}. 

A common approach to measuring psychological aspects in text is to validate dictionaries: curated sets of words, with subsets aimed at measuring each component of the psychological aspect in question \cite{pennebaker2015development, graham2009liberals, holtrop2022exploring, ponizovskiy2020development}. Some work estimating the values of the authors of text has been conducted on individuals who have written personal essays and social media posts e.g.\ \cite{maheshwari2017societal, ponizovskiy2020development}, and in arguments abstracted from various forms of public facing text \cite{kiesel2022identifying}. However, we have not seen work aimed at measuring values \textit{perceived} in text, measuring them along a scale as in prior work \cite{schwartz1992universals}, or ultimately treating them as a hierarchical list in line with theory \cite{rokeach1973nature}. 

\section{Primary Lyrics Data}

We aim to collect a sample of lyric data where the lyrics are as accurate as possible, and our sample is as representative as possible. We sampled from the population of songs in the Million Playlist Dataset (MPD)\footnote{\url{https://research.atspotify.com/2020/09/the-million-playlist-dataset-remastered/}} as it is large and recent compared to other similar datasets. The lyrics themselves were obtained through the API of \href{https://www.musixmatch.com/}{Musixmatch}\footnote{\url{https://www.musixmatch.com/}}, a lyrics and music language platform. Musixmatch lyrics are crowdsourced by users who add, correct, sync, and translate them. Musixmatch then engages in several steps to verify quality of content, including spam detection, formatting, spelling and translation checking, as well as manual verification by over 2000 community curators, and a local team of Musixmatch editors. Via their API, Musixmatch provided us with an estimated first 30\% of the lyrics of each song. 

Using the `fuzzy' stratified sampling method described below, we sampled 2200 songs. Three members of the research team manually screened approximately 600 of the 2200 songs for inclusion. Each set of lyrics was confirmed to be a match to the actual song, and for suitability\footnote{Each member independently screened each lyric and the screening process overall was discussed at length.}. Lyrics were unsuitable if they were: 1) not English, 2) completely onomatopoetic, 3) repetitions of single words or phrases, 4) too few words to estimate values present or, 5) were not a match to the meta-data of the song, e.g. artist title, song name. This resulted in 380 songs, 20 of which were used in a pilot study to determine the number of ratings to gather per song, and 360 were used for annotation. 

\subsection{Fuzzy Stratified Sampling}\label{sec:primarylyricdata:fuzzysampling}

An initial challenge is determining how to represent a corpus. In our case, the population of songs is known to be very large\footnote{e.g., Spotify reports over 100 million songs in its catalogue\url{https://newsroom.spotify.com/company-info/}}. An ideal scenario would be one in which we aim for a known number of songs, randomly sampled from within clearly defined strata, i.e. relevant subgroups, also known as \textit{stratified random sampling}~\cite{groves2009survey}. However, for music, we do not know how many songs we would need to sample in order to reach saturation, what the relevant strata to randomly sample within should be, and how to measure relevant parameters from each stratum. 

Some measurable strata that affect the use of language in song lyrics are clear: e.g., the year of release, which may reflect different events or time-specific colloquial slang. Others are less clear: e.g., there is no single metric of popularity for music, although it can be estimated from various sources such as hit charts. Some may be very subjective, such as genre, for which there may be some overlap of human labelling, but no clear taxonomy exists in the eyes of musicological domain experts~\cite{Liem2012MusicGap}. 

Based upon these considerations, we aimed for a stratified random sampling procedure, based on strata that we acknowledge to be justifiable given our purpose, yet in some cases conceptually `fuzzy': (1) release date; (2) popularity, operationalized as artist playlist frequency from the MPD~\cite{DBLP:conf/recsys/ChenLSZ18}; (3) genre,  estimated from topic modeling on Million Song Dataset artist tags  \cite{schindler2012facilitating}; (4) lyric topic, through a bag-of-words representation of the lyrics data. Popularity and Release date were divided into equally spaced bins; e.g. we divided release year into decades (60s, 70s, 80s, and so on), and genre and lyric topic were divided into categories. 

Release date was quantized into 14 bins in 10-year increments from 1890-2030. Popularity was exponentially distributed, and thus manually binned, to make the quantiles per each of the 7 bins as similar as possible. Thus, the first bin contained the lowest 40\% of the population in terms of popularity, while the 7th bin contained the highest 9\%. Topic modelling was applied on a bag-of-word representation of the lyrics data and artist-tag data to yield $25$ estimated genres and $9$ lyrics topic strata, respectively.

We observed a skewness of data concentration with regard to several of our strata, e.g., songs that are recent and widely popular are most likely be drawn. To correct for this and thus get a more representative sample of an overall song catalogue, we oversample from less populated bins. For this, we use the maximum-a-posteriori (MAP) estimate of the categorical distribution of each stratum. The oversampling is controlled by concentration parameter $a$ of the symmetric Dirichlet distribution. We heuristically set this parameter such that songs in underpopulated bins still will make up up 5-10 \% of our overall pool\footnote{Full code of our sampling procedure is at~\url{https://anonymous.4open.science/r/lyrics-value-estimators-CE33/1_stimulus_sampling/stratified_sampling.py}}. Through this method, we subsampled our initial 2200 songs lyrics.

\begin{figure}
 \centerline{\framebox{
 \includegraphics[width=1\columnwidth]{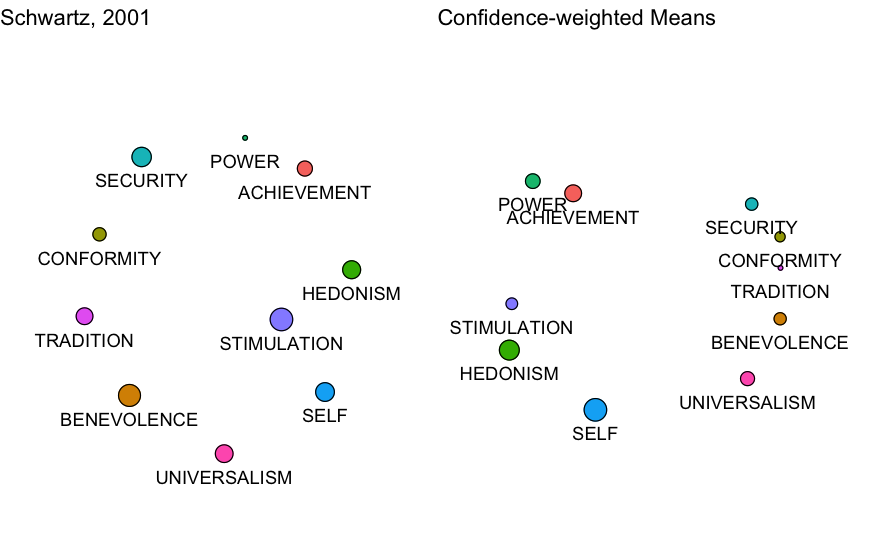}}}
 \caption{MDS plots derived from the correlation plot reported in \cite{schwartz2001extending}, and our participant responses as confidence-weighted means}
 \label{fig:mds_plots}
\end{figure}

\section{Ground-Truthing Procedure}\label{sec:groundtruthing}

We chose to obtain our annotations from samples of the US population, representative in terms of self-reported sex, ethnicity and age, through the Prolific\footnote{\url{https://prolific.co}} platform. Annotator pools comprised of two samples, the first n=505 wave participated in a pilot study to estimate the number of ratings per song needed on average, and the second n=600 wave comprised our main data collection. Participants completed the survey on the Qualtrics \footnote{\url{https://qualtrics.com}} platform. 

We clearly differentiate between the Author and the Speaker of lyrics by explaining to participants that the Author of song lyrics may write from the perspective of someone or something else (the Speaker). 17 randomly selected sets of lyrics were then shown to each participant along with instructions to annotate each with the values of the Speaker. We adapted the 10-item questionnaire used in \cite{lindeman2005measuring} for the value annotations, as it is the shortest questionnaire for assessing personal values whose validity and reliability have been assessed\footnote{It has shown correlations ranging from .45-.70 per value with longer more established procedures, test-retest reliability, as well as the typical values structure shown in Figure~\ref{fig:circle}}. As in \cite{lindeman2005measuring}, each questionnaire item is a specific value along with additional descriptive words e.g. POWER (social power, authority, wealth). We adjusted it by asking participants to indicate the values of the Speaker of the lyrics, and by having them indicate on a bar with -100 (opposed to their principles) on one end, and +100 (of supreme importance) on the other end instead of a likert scale. In addition, we asked participants to indicate how confident they were in their ratings, on a scale of 0 (not at all confident) to 100 (extremely confident), inspired by work that has shown that self-reported confidence in ratings can be used to estimate the accuracy of individual ratings \cite{cabitza2020if}. 

We used a procedure similar to \cite{DeBruine_Jones_2018} in order to determine the number of raters. Specifically, we recruited a representative 500+ participant sample of the US using the Prolific platform, who completed our survey for 20 songs. We then computed canonical mean ratings of each of the 10 values per song, and inter-rater reliability using Cronbach's Alpha. We then estimated Cronbach’s alpha for a range of subsample sizes (5 to 50 participants in increments of 5), for each of the 10 values. We repeated this procedure 10 times per increment, separately for each of the 10 values, and examined the distribution of Cronbach’s Alpha. We specifically looked for the sample size with which Alpha exceeded .7 \footnote{.7 is a commonly considered an acceptable level of reliability in the form of internal consistency}. We arrived at a conservative estimate of 25 ratings per set of lyrics, with songs receiving a median 27 ratings (range 22-30). 

\subsection{Reliability, Agreement and Initial Validation}

The rater reliability was estimated via intra-class correlation for each personal value, (type 2k: see \cite{koo2016guideline}) using the `\texttt{psych}' package in R \cite{Psychcitation}, all of which exceeded .9 (excellent reliability). As an initial validation, we compare data simulated from values in the upper triangle of a correlation matrix reported in~\cite{schwartz2001extending} to those derived from our study. To aggregate our participants rankings for this purpose, we compute confidence-weighted means inspired by \cite{cabitza2020if}: we estimate confidence-weights by dividing participant’s self-reported confidence of a given rating by the highest possible response (100), and then compute aggregated means weighted by these. For both the simulated data and confidence-weighted mean scores, we generate a multi-dimensional scaling plot (MDS) \cite{davison2000multidimensional} for visual comparison, which has previously been used as method to assess measurements conform to theory~\cite{ponizovskiy2020development, lindeman2005measuring}. Note: the interpretation is to observe whether each of the values appears next to expected neighboring values, and not each value's orientation. From these plots (Figure~\ref{fig:mds_plots}), in as little as our 360 annotated lyrics, we surprisingly see similar clusters and relative positioning relations emerging as those obtained from a formal cross-cultural study. 

We coerced the annotated scores to ranked lists of values, such that the highest scoring value was at the top. We derived ranked lists per participant per song, and then used Robust Ranking Aggregation (RRA) to extract a single ranked list per song. Aggregation was conducted using R version 4.2.2.\cite{Rcitation}, and the \texttt{RobustRankAggreg} package \cite{RRAcitation}. Briefly, RRA produces a ranked list by comparing the probability of the observed ranking of items to rankings from a uniform distribution. Essentially, scores are determined by comparing the height of an item on a set of lists to where it would appear if its rank were randomly distributed across lists. These scores are then subjected to statistical tests, where the resulting \textit{p} value is Bonferronni corrected by the number of input lists\cite{kolde2012robust}. Thus, when an item appears in different positions on a list, the resulting p value is high, as its position appears randomly distributed. 

As lyrics are ambiguous, we expect that some songs' values are completely subjective. We operationalize these as randomly distributed rankings for all personal values for completely subjective songs, i.e. \textit{p} values above .05 for all 10 items on the ranked list. Results from the RRA show 62 songs with \textit{p} values above .05 for all 10 values, and 96 songs with only 1 value ranked. At most, 5 values were ranked, which occurred for 35 songs. Thus, we confirm that although there was correspondence in the scores that participants assigned per value per song, ranked lists did not always agree. 

\begin{figure}
 \centerline{\framebox{
\includegraphics[width=1\columnwidth]{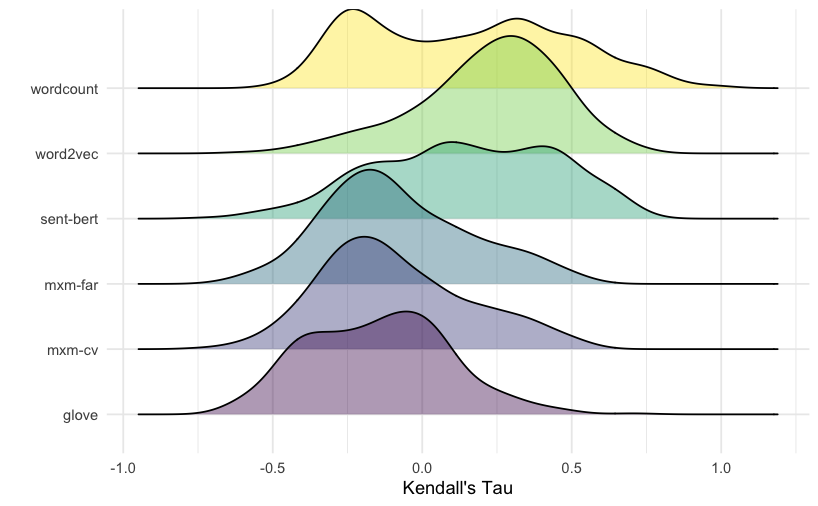}}}
 \caption{Rank correlations between NLP systems / word counts and Robust Ranking Aggregation lists, by normalization scheme.}
 \label{fig:rank_corr}
\end{figure}

\section{Automated Scoring}
For automated scoring, we use a dictionary of words associated with the 10 Schwartz values~\cite{ponizovskiy2020development}. With this dictionary as reference, we computationally estimate the degree to which each value is reflected in the lyrics text according to traditional word counting~\cite{ponizovskiy2020development}, as well as by assessing cosine similarity between dictionary words and lyrics texts using four classes of pre-trained word embeddings: \texttt{word2vec}, a generic English word embedding trained on Google News dataset~\cite{DBLP:conf/nips/MikolovSCCD13}; \texttt{glove}, another generic English word embedding trained on Common Crawl dataset~\cite{DBLP:conf/emnlp/PenningtonSM14}; \texttt{mxm-far-[1$\sim$10]}, trained on the collected initial lyrics candidate pool, employing the Glove model~\cite{DBLP:conf/emnlp/PenningtonSM14} (using ten models populated from ten cross-validation folds, whose parameters are tuned based on English word similarity judgement data~\cite{DBLP:conf/acl/FaruquiD14}.); \texttt{mxm-cv-[1$\sim$10]}, ten variants of lyrics based word-embeddings from cross-validation folds selected by Glove loss values on the validation set; and finally, \texttt{sent-bert}, a transformer model that encodes sentence into a embedding vector, fine-tuning of a generic self-supervised language model called MPNet, which is trained on a large scale English corpus~\cite{DBLP:conf/emnlp/ReimersG19}. Our process thus resulted in 24 sets of scores: 5 from models and one from word-counting, normalized using four methods. 

We take the perspective from theory that that value assessments should be seen as ranked lists, and thus coerce scores to ranked lists per model per song. We then compute rank correlations between ranked lists derived from model scores and RRA lists from participants. As RRA lists assess lack of consensus on rankings, personal values with high \textit{p} values received tied rankings, at the bottom of the list. Correlations were computed using Kendall's $\tau$ which is robust to ties (Figure~\ref{fig:rank_corr}).  

In earlier work~\cite{richard2003one, ponizovskiy2020development}, Pearson correlations of 0.1-0.2 were considered as moderate evidence of the validity of a proposed dictionary in relation to a psychometrically validated instrument. Although we are using a different metric, we observe several models whose mean rank correlations exceed the .10 mark. The mean Kendall's $\tau$ values were highest for the word2vec, sent-bert, and wordcount models with null normalization (SD=.24, .30, and .34 respectively). We further observe that 76\%\ of the rank correlations for word2vec exceed the .10 mark, followed by 56.1\%\ from sent-bert, and 47.8\%\ from wordcounts. Although none of these models had been thoroughly optimized and thus this cannot be interpreted as a thorough benchmark, we do see evidence of higher than expected correlations. 

\begin{figure}
 \centerline{\framebox{
\includegraphics[width=1\columnwidth]{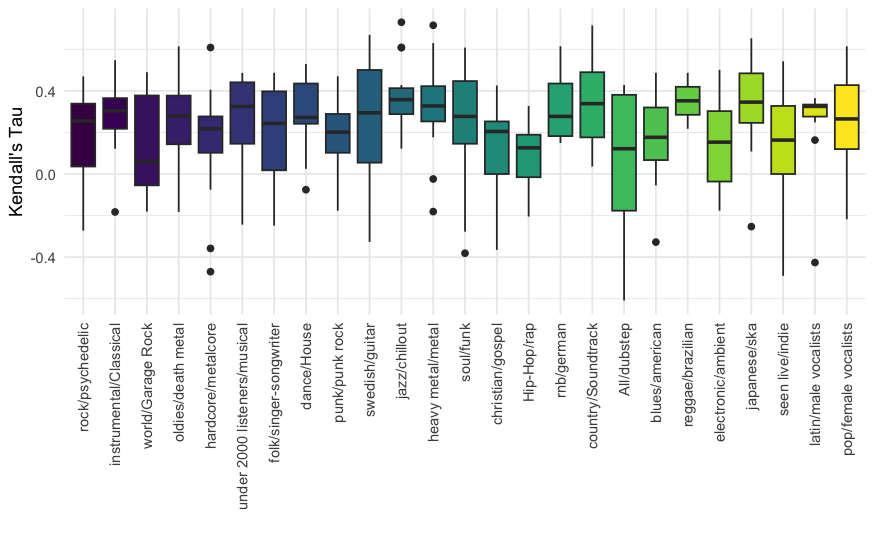}}}
 \caption{Rank correlations between word2vec scores Robust Ranking Aggregation lists, per genre grouping operationalized as Artist Tag Topic.}
 \label{fig:artist_tag_topic}
\end{figure}

We also explored whether our fuzzy strata might hint towards more or less automatically scorable lyrics. We found most strata to be uninformative. However, when examining the rank correlations for our overall best performing model, word2vec, we did observe higher mean correlations for some artist tag topics than others (Figure~\ref{fig:artist_tag_topic}). In particular, topics 10 (which included the tags: ‘jazz’, ‘chillout’, ‘lounge’, ‘trip-hop’, ‘downtempo’), 11 (which included the tags like: ‘metal’, ‘celtic’, ‘thrash metal’, ‘dutch’, ‘seen live’), and 16 (which included tags like: 'country’, ‘Soundtrack’, ‘americana’, ‘danish’, ‘Disney’). Although speculative, we do expect that certain genres are more difficult to interpret than others, in particular for people who are generally unfamiliar with such music.  

\section{Descriptive Analyses}

We conduct a further exploratory data analysis by examining the gathered value annotations with respect to the song strata introduced in Section~\ref{sec:primarylyricdata:fuzzysampling}. To better understand the overall patterns of value rankings in songs we visualize the average ranking of each value for each level of each stratum. To reflect the uncertainty of aggregated ranking from RRA, we employ `truncated' rankings: the values within each aggregated ranked list are considered ties if their p-values higher than the threshold ($p=0.05$), hence with high uncertainty in their ranking positions.\footnote{We assume that the adjusted exact p-value from RRA monotonically decreases as the rank position ascends (i.e., the lower the p-value is, the higher the ranking position is).}


In all results, we observe that there is a tendency of overall value ranking: 1) a generally strong presence of HEDONISM in higher ranks in all cases, followed by STIMULATION and SELF (SELF-DIRECTION). 2) ACHIEVEMENT and POWER generally follow next across all figures, and 3) the rest of the values, including BENEVOLENCE, UNIVERSALISM, SECURITY, CONFORMITY, and TRADITION overall rank lower, but show higher variability across strata. We refer to these three groups of values as \emph{Group1} (HEDONISM, STIMULATION and SELF), \emph{Group2} (ACHIEVEMENT and POWER), and \emph{Group3} (the rest) for the rest of the section.

\begin{figure}
 \centerline{\framebox{
\includegraphics[width=1\columnwidth]{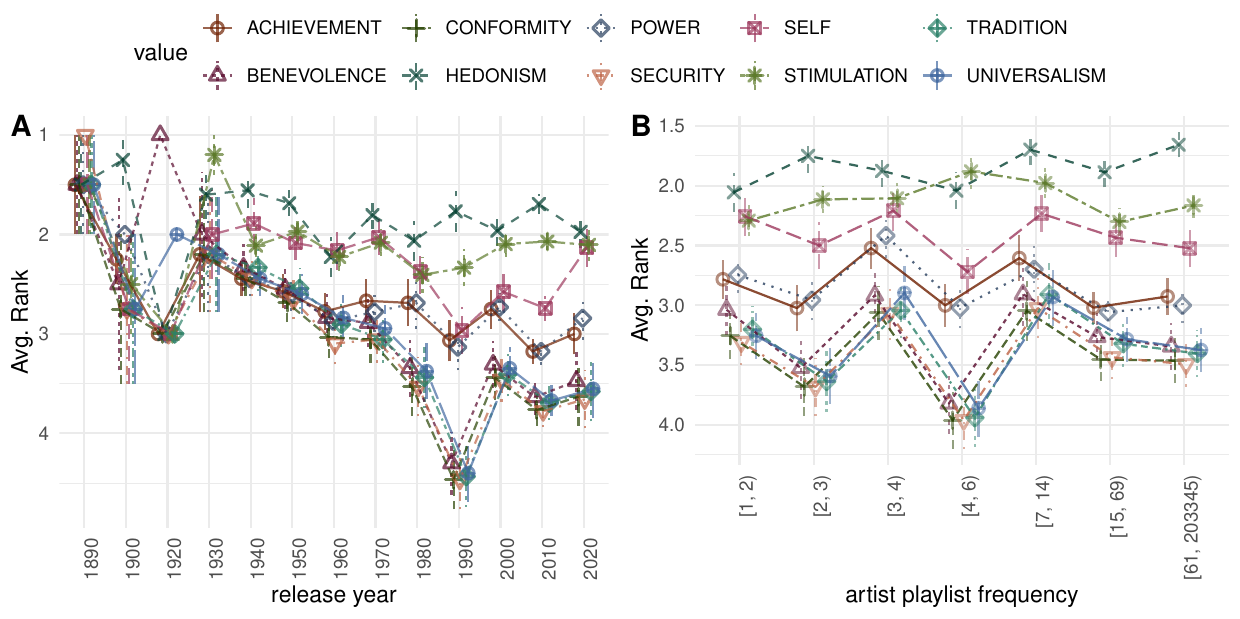}}}
 \caption{Average value ranking from `release year' (A) and `artist-playlist frequency' (B). $x$ and $y$ axis represent the strata and average ranking measure from RRA, respectively.
 Each point in different point shapes and vertical bars denote the average ranking value and its confidence interval (at 95\% level). For visual convenience, we connected the same values with lines.
 }
 \label{fig:avg_rank_trend_12}
\end{figure}

Zooming in each to stratum, in Figure \ref{fig:avg_rank_trend_12}, we observe that the `release year' (sub-figure \textbf{A}) strata show the most consistent and visible trend especially for Group3, which generally declines over time. Such a trend is not as obvious in Group1 and only partially observed in Group2. The low presence of Group3 is especially noticeable in the 1990s, although it regained its presence to some degree, a pattern which the SELF value from Group1 partially shares. Such visible movements suggest that the rank of specific values may evolve over time. In sub-figure \textbf{B}, we observe the most flat response across all strata considered: beyond the fluctuation pattern that is shared by all groups, there is no substantial variability among groups, which implies that popularity might not be as correlated as the `release year'.


\begin{figure}
 \centerline{\framebox{
\includegraphics[width=1\columnwidth]{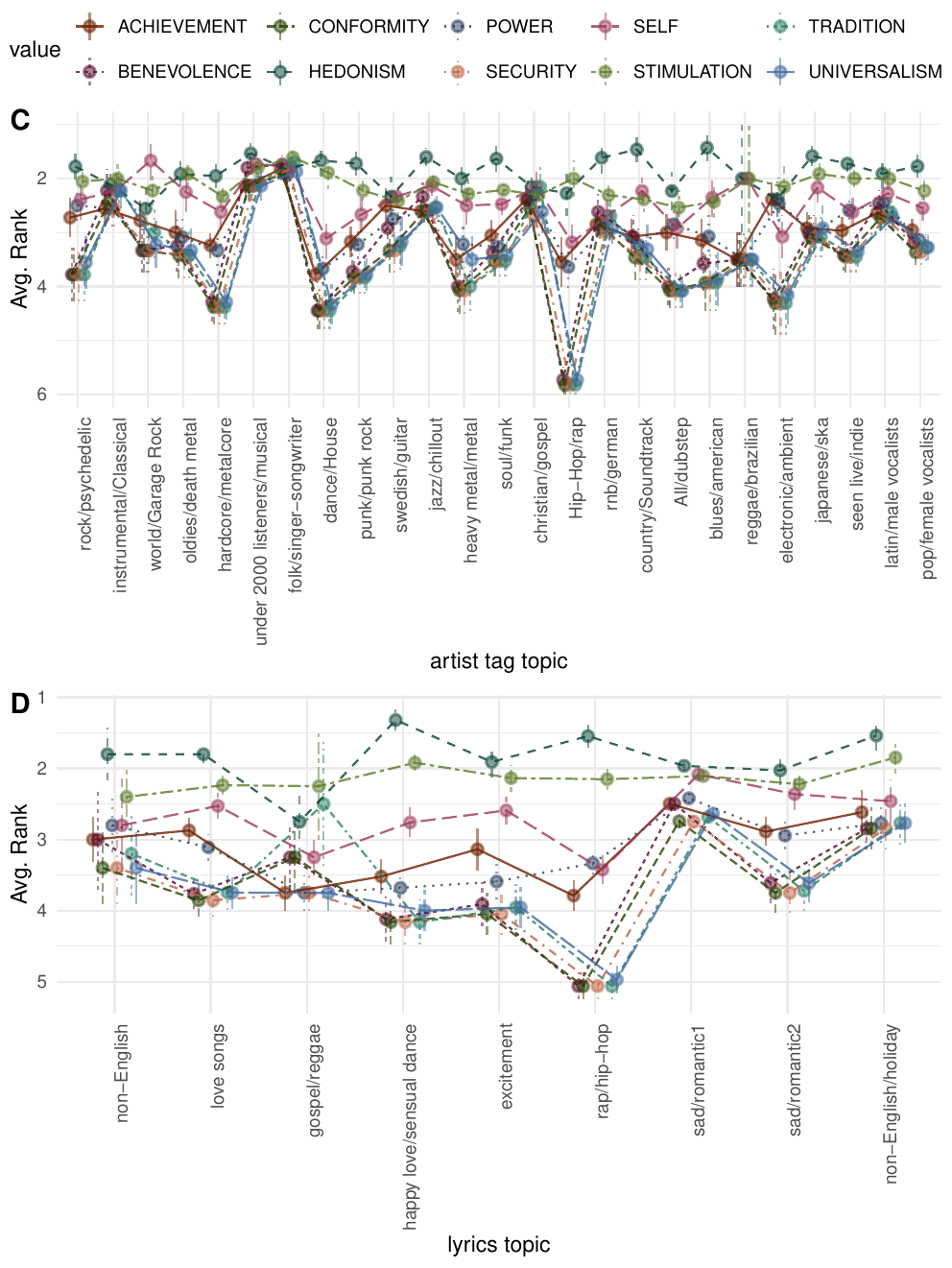}}}
 \caption{Average value ranking from `artist-tag topic' (C) and `lyrics topic' (D).}
 \label{fig:avg_rank_trend_34}
\end{figure}


Moving onto Figure~\ref{fig:avg_rank_trend_34}, we discuss the value presence pattern in two `topic' strata. First, in sub-figure \textbf{C}, we observe that Group3 values show overall higher variability than `artist playlist frequency'. It is notable that there are a few distinct topics in which Group3 values show a significant difference; the sixth, seventh and fourteenth topics, which correspond to the `under 2000 listeners/musical', `folk/singer-songwriter', and `Hip-Hop/rap' topics when represented in primary topic terms. Specifically, we see that first two topics show a high presence of Group3 values, while the latter topics show the least presence of Group3 values. It suggests that the artists in these styles/genres were perceived on average to present clearly different sets of values through their lyrics, distinguished by the inclusion/exclusion of values such as BENEVOLENCE or UNIVERSALISM.


Finally, considering sub-figure \textbf{D}, we observe a similar pattern as `artist playlist frequency' in \ref{fig:avg_rank_trend_12}, albeit with relatively more variability in Group3 values. Notably, the `rap/hip-hop' lyrics topic shows the least presence of Group3 values, which aligns to the observation from previous sub-figure. The `sad/romantic1' topic, on the other hand, shows the highest ranking of Group3 values. Another remarkable topic is `gospel/reggae' topic, where HEDONISM value is least present, which semantically aligns well with the typical lyrical theme of those songs.

\section{Limitations and Future Work} 
In this work we attempt to ground-truth perceptions of ambiguous song lyrics for perceived human values. We adopt a validated questionnaire from the social sciences for this purpose, in addition to a purposeful, if conceptually 'fuzzy', stratified sampling strategy, and estimate the average number of ratings needed to estimate the average perception of values in a song. We acknowledge our current sample of 360 lyrics is small and may need expansion for more typical work, and that, while we had a representative population sample, not every member of the sample rated every song. We thus did gather diverse opinions, but cannot claim they fully represent the target population. In addition, the small sample of songs allowed for only limited observation of patterns that might emerge in larger samples with relation to our defined strata, and indefinite conclusions given the overall massive population of songs in existence. We also did not assess whether variations on the annotation instrument might result in substantial differences in the annotations we received \cite{kern2023annotation}, nor did we repeat our procedure \cite{inel2023collect}. In addition, we acknowledge that participants from different groups will perceive and thus annotate corpora differently~\cite{homan2022annotator, prabhakaran2023framework}. Thus, we expect that lyrics may be especially sensitive to varying perceptions, which we did not explore in this work. Finally, we only provide a preliminary comparison to automated scoring methods, and did not leverage the most contemporary tools for this purpose (e.g. Large Language Models). All of these are rich and promising avenues for future work.  

The most interesting avenues are potential relationships that could be revealed with more annotated songs, and eventual automated scoring methods. In particular, we see potential in understanding music consumption more broadly from patterns revealed in the dominant value hierarchies in specific music genres, popularity segments, lyrical topics, and even release year. And for understanding music consumption more narrowly, from patterns revealed in an individual's music preferences, and the degree to which they conform with their own value hierarchy. 

\section{Conclusion}

Song lyrics remain a widely and repeatedly consumed, yet ambiguous form of text, and thus a promising and challenging avenue for research into better understanding the people that consume them. We observe promising initial results for the annotation of personal values in songs, despite our limitations. MDS plots of aggregated ratings showed the beginnings of the expected structure of values, conforming more closely than might be expected from as little as 360 songs. We also observed high inter-rater reliability in the raw scores, suggesting a sufficiently reliable annotation procedure with 25 ratings. Thus, we see promise on our method for ground-truthing lyrics despite their ambiguity. A post-hoc procedure revealed that 15 ratings may be enough on average: we repeatedly subsampled 5, 10, 15 and 20 ratings for each value within each song, and calculated pearson correlations between subsample means and canonical means. From this, we see Pearson correlations to the canonical mean exceed 0.9 for all values from 15 subsampled ratings. Further lyric annotation may thus require fewer annotations per song than what was gathered in this work. In addition, we observe promising rank correlations between ranked rater scores and our automated methods, with over 75\%\ of the rankings in our best performing model above a minimal threshold of .10. Despite inherent challenges in the task, our method shows initial promise, and multiple fruitful avenues for future work. 

\section{Ethics Statement}

Our study includes data gathered from people, and was approved by the Human Research Ethics board of our university. We follow Prolific's guidelines on fair compensation to set our compensation rates. Survey design and data handling were pre-discussed with our institutional data management and research ethics advisors, we obtained formal data management plan and human research ethics approval. Participants gave informed consent before proceeding with the survey, which informed them of the intentions of use for their data, and that it could be withdrawn at any time. 

\bibliography{ISMIR2024_template}

%
%
%
%
%

\end{document}